\documentclass[final,5p,times,twocolumn]{elsarticle}
\usepackage{lscape}
\usepackage{url}
\usepackage{booktabs} % For formal tables
\usepackage{caption} % For formal tables
\usepackage{array}    % For table and array extensions

\usepackage[labelfont=bf]{caption}
\usepackage{tabularx}
\usepackage{amssymb}
\usepackage{amsthm}
\usepackage{amsmath}

\journal{arXiv}
\begin{document}

\begin{frontmatter}



\title{Forecasting Day-Ahead Electricity Prices in the Integrated Single Electricity Market: Addressing Volatility with Comparative Machine Learning Methods}

\author{Ben Harkin}
\author[2]{Xueqin Liu\corref{cor1}}

\affiliation[2]{organization={School of Electronics, Electrical Engineering and Computer Science, Queen's University Belfast},
            city={Belfast},
            postcode={BT9 5AH}, country = {UK}, email={, x.liu@qub.ac.uk}}
            
\cortext[cor1]{Corresponding Author: Xueqin Liu - x.liu@qub.ac.uk}

\begin{abstract}
This paper undertakes a comprehensive investigation of electricity price forecasting methods, focused on the Irish Integrated Single Electricity Market, particularly on changes
during recent periods of high volatility. The primary objective of this research is to evaluate
and compare the performance of various forecasting models, ranging from traditional machine
learning models to more complex neural networks, as well as the impact of different lengths of
training periods.
The performance metrics, mean absolute error, root mean square error, and relative mean absolute error, are utilized to assess and
compare the accuracy of each model. A comprehensive set of input features was investigated and selected from data recorded between October 2018 and September 2022. 
The paper demonstrates that the daily EU Natural Gas price is a more useful feature for electricity price forecasting in Ireland than the daily Henry Hub Natural Gas price. 
This study also shows that the correlation of features to
the day-ahead market price has changed in recent years.
The price of natural gas on the day and the amount
of wind energy on the grid that hour are significantly more
important than any other features. More specifically speaking, the input fuel for electricity has become a more important driver of the price of it, than the total generation or demand.  In addition, it can be seen that System Non-Synchronous Penetration (SNSP) is highly correlated with the day-ahead market price, and that renewables are pushing down the price of electricity. 

\end{abstract}


\end{frontmatter}


\section{Introduction}
\label{sec:intro}

%%       \citet{<label>} ==> Jones et al. [21]
%%       \citep{<label>} ==> [21]
%%

The electricity market in Ireland has evolved rapidly in recent years, with the impact of renewable energy sources and the recent rise in natural gas prices leading to volatile prices. These tumultuous times have left the electricity sector with the critical challenge of accurate forecasting of electricity prices, which is crucial for effective decision-making. The Integrated Single Electricity Market (I-SEM) is the wholesale electricity market that facilitates the trading of electricity on the island of Ireland. The I-SEM operates on a day-ahead and intra-day basis, as well as having balancing markets and capacity markets, with a variety of market participants, including generators, suppliers, and traders. Accurate electricity price forecasting is essential for market participants to make informed decisions about production, consumption, and trading activities.
The study uses historical electricity price data from the I-SEM and other relevant sources to develop the forecasting models. The impact of various factors on forecasting performance is also studied, including demand, supply and weather conditions, and varying lengths of training data. The performance of the models are evaluated using various accuracy metrics, including mean absolute error, root mean square error, and relative mean absolute error.
This paper makes a significant contribution to the field of electricity price forecasting in Ireland:
\begin{itemize}
  \item Analysing what data features are most correlated to the price of electricity and how that has evolved in  recent years. For example, for the first time, the EU Daily Natural Gas price was used instead of the daily Henry Hub Natural Gas Price.
  \item Understanding what models deliver the most accurate results, and how that has changed.
  \item Understanding what data features and training data lengths deliver the most accurate results, and how that has changed.
  \item Delivering a thorough study of research on electricity price forecasting for the I-SEM, incorporating a comprehensive dataset covering the volatile period from October 2018 to September 2022.
  \item The use of a new metric, Relative Mean Absolute Error, which delivers useful and insightful findings for analysing periods with significant price changes.
 
\end{itemize}
The findings of this research are intended to benefit a range of stakeholders, including market participants (such as generators and retailers), policymakers, and fellow researchers. While the primary focus is on Ireland, the insights gained may also be applicable to similar electricity markets worldwide.
\subsection{Overview of the Irish Single Electricity Market}
\label{sec:overview}
The Integrated Single Electricity Market was brought online on the 30th of September 2018. It is designed as an improvement on the previous market, that combined the Northern Ireland and Republic of Ireland systems in 2007 \citep{SEM1}. It is designed with the explicit aim of delivering “increased levels of competition which should help put a downward pressure on prices as well as encouraging greater levels of security of supply and transparency” \citep{Eir1}. 
It is operated by the Single Electricity Market Operator (SEMO).

\begin{figure}[h]
    \centering
    \includegraphics[width=1\linewidth]{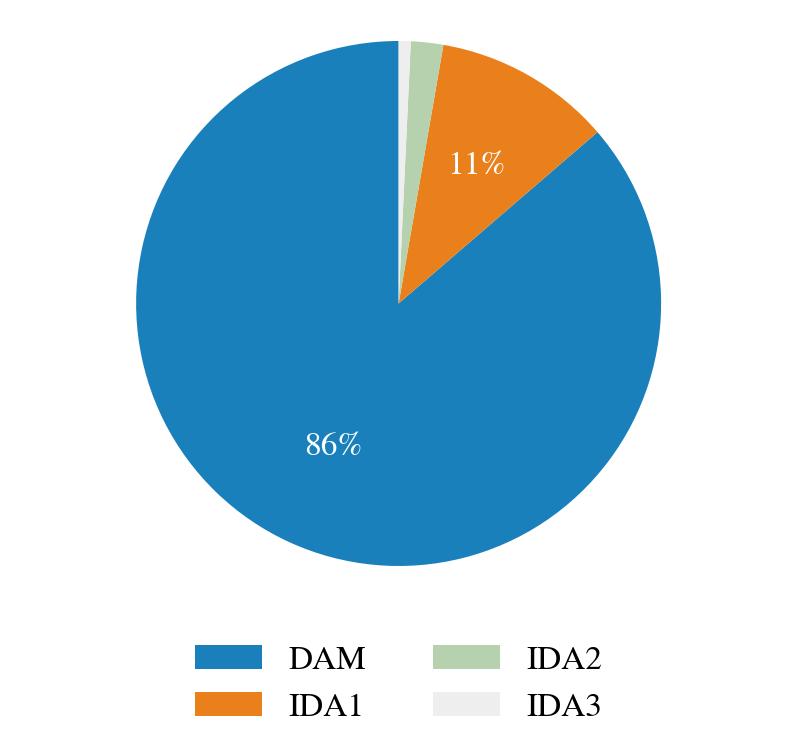}
    \caption[Market share of different I-SEM markets, Oct 2021 - Sept 2022]{Market share of different I-SEM markets, Oct 2021 - Sept 2022 ~\cite{SEM2}}
    \label{fig:marketshare}
\end{figure}

Different markets are used to ensure there is a balance between electricity demand and electricity generation (also known as load). One of these is the day-ahead market. As can be seen in Figure \ref{fig:marketshare},the day-ahead market is the market where most of the electricity trading is done, it made up over 86 per cent of the ex-ante market share in the period October 2021 until September 2022 \citep{SEM2}.

\begin{figure}
    \centering
    \includegraphics[width=1\linewidth]{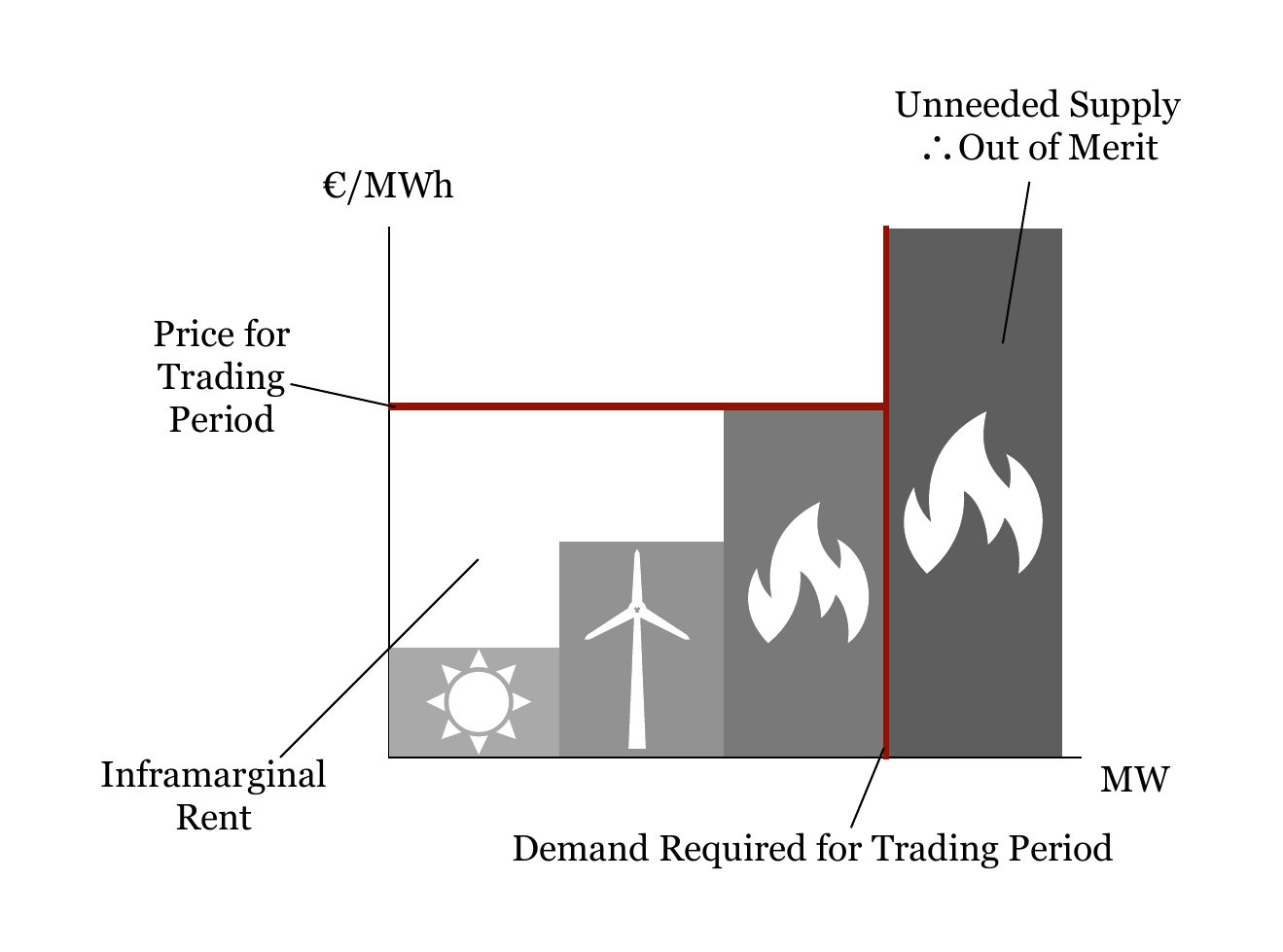}
    \caption{Diagram of how the day-ahead price of electricity is set.}
    \label{fig:DAMPriceSet}
\end{figure}

This dominance of the trading portion makes it an ideal market to focus on, as improved efficiency will have a far greater impact on market participants compared to the other markets. Bids and offers can be submitted from 19 days before the market closes, which is at 11:00 the day before delivery. SEMO release the prices for the 24 hours (23:00-23:00) of the day of interest at 13:00 the day before. 
Generators bid to supply a certain amount of electricity for each one-hour trading period for a certain price. The cheapest of these that fulfil the necessary demand are chosen, with the price being set by the most expensive generator used. Those generators that were not used are deemed ‘Out of Merit’. The difference between the price submitted by the generator and the price they receive is known as ‘inframarginal rent’. It can be seen in Figure \ref{fig:DAMPriceSet}, how cheap renewables, push out more expensive generators, setting the price at a lower value than it would have been.

\begin{figure}
    \centering
    \includegraphics[width=1\linewidth]{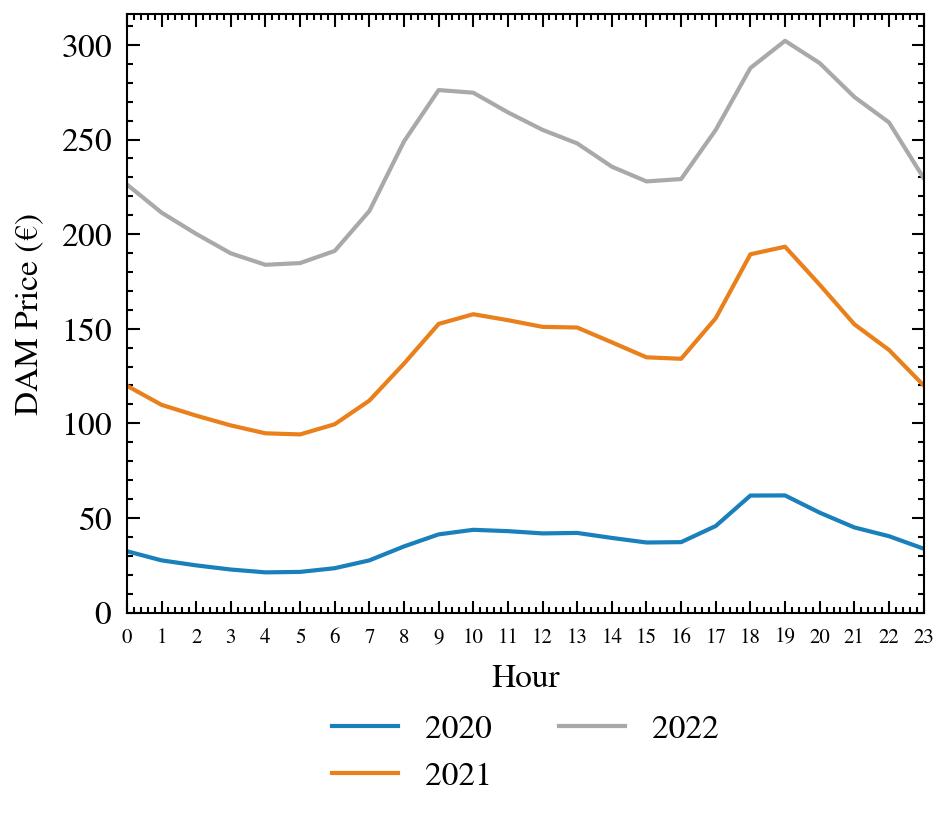}
    \caption{Average daily price, 2020 - 2022}
    \label{fig:dailyaverage}
\end{figure}

The price of electricity is impacted by a multitude of factors, including supply, demand, and input fuel costs, which in turn are affected by a variety of market forces, like global weather fluctuations and geopolitical events \citep{Su2023}. Electricity markets are unusual in that electricity cannot be stored (at least at scale), this leads to electricity markets having sudden spikes or declines. The price of electricity is also affected by calendar indices like hour of the day and day of the week.
The market has changed in recent years. There was a fall in prices in 2020, which was seen around the globe, and caused by lower electricity demand due to the Covid-19 pandemic and lower costs for natural gas (due to the fall in electricity and industrial demand) \citep{IEA2020} \citep{Ghiani2020}. Ireland also had an increase in wind energy compared to 2019, further pushing down prices \citep{Mohamed2022}.
The day-ahead market price has risen in recent years and become more tumultuous. This is not limited to Ireland but is replicated in European markets also \citep{Tschora2022}. This rise in prices can be seen in Table 1, with the rise in standard deviation after 2020 reflecting the increasingly tumultuous nature of the market. Figure \ref{fig:dailyaverage} clearly shows the rise in the average price for each hour.

\begin{table}[ht]
  \label{tab:annualbreakdown}
  \captionsetup{labelsep=newline,justification=justified,singlelinecheck=false}
  \caption{Analysis of the day-ahead market from 2019 - 30th Sept 2022.}
  
  {\scriptsize
  \begin{tabular}{lcccc}
    \toprule
    Year & 2019 &  2020 & 2021 & 2022 (1st Jan – 30th Sept) \\
    \midrule
    Mean & 50.26 &37.63 & 136.07 & 239.83\\
    Standard Deviation &23.68	&23.05	&80.39	&105.22\\
    Minimum Value &	-11.86&	-41.09	&-20.43&	-25.78 \\
    Maximum Value 	&365.04&	378.12&	500&	705.47 \\
    \bottomrule
  \end{tabular}
  }
\end{table}

\begin{figure}
    \centering
    \includegraphics[width=1\linewidth]{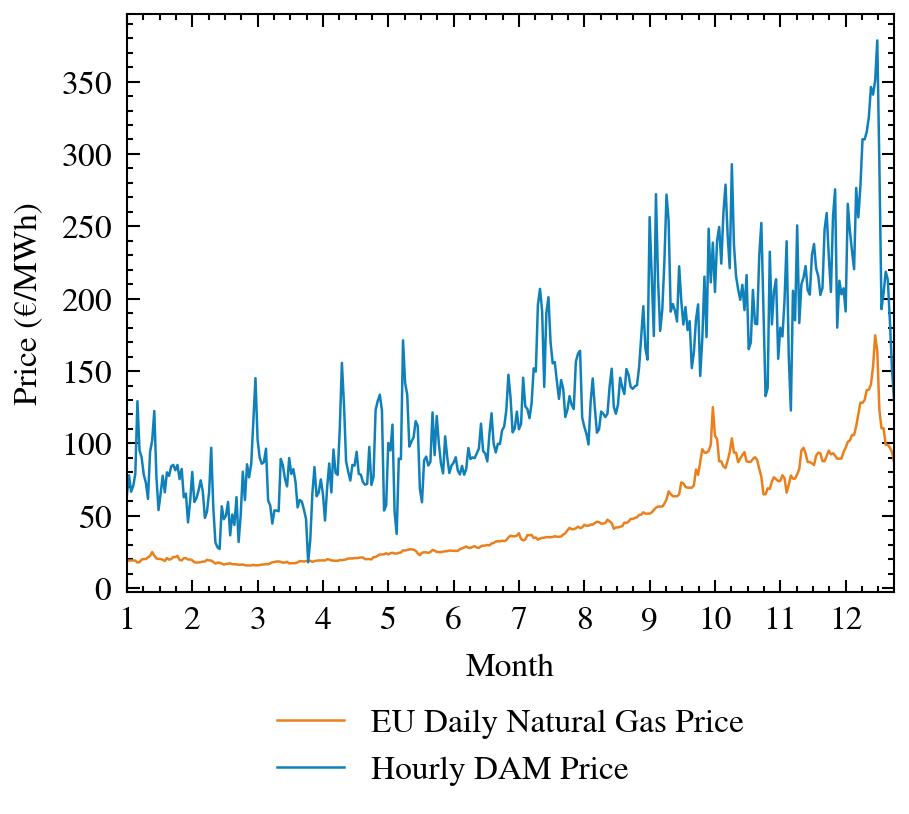}
    \caption[The DAM Electricity Price compared to the Natural Gas Price, 2021]{The DAM Electricity Price compared to the Natural Gas Price, 2021 ~\cite{TE2023}}

    \label{fig:pricevgas}
\end{figure}

Looking closer at the price increase over 2021 in Figure \ref{fig:pricevgas}, it can be seen how the price rises with the rise of the price in natural gas.

\subsubsection{Electricity Price Forecasting}
\label{sec:overview_priceforecasting}

Electricity price forecasting can be used by generators, suppliers or users of energy to estimate prices before they occur. The most important aspect is that accurate price forecasting allows generators and suppliers to act more efficiently than they would without it \citep{Croonenbroeck2017} \citep{Ugurlu2018-hydro-based}. Just as we know that more competition helps bring down prices \citep{Amenta2022} \citep{Andrews1}, it can be understood that accurate forecasting allows for more and better, more efficient competition in the markets, pushing down prices for consumers. 
Building electricity management systems, for home use or for industries like agriculture, that use electricity price forecasts to schedule tasks throughout the day, lead to “significant cost savings, as compared to cost-unaware or day-night-tariff based schedules” \citep{Grimes2014}. Accurate price forecasting also allows for better demand scheduling for industries with high electricity requirement like data centres \citep{Albahli2020}. More accurate prediction models could also be used to create “early-warning systems” that warn generators, suppliers and citizens, days in advance when electricity prices are forecast to be high, so that demand can be reduced \citep{Alshater2022}. 
As can be seen in Table 1, volatility has increased in recent years, this would mean that the price of electricity has become more difficult to forecast. Research from the UK Department of Business, Energy \& Industrial Strategy finds that in their models of the future, “wholesale prices will become increasingly volatile… periods of very low prices and very high prices, and less periods between these extremes” \citep{Dept1}. The similarity between the UK and Irish markets allows for similar conclusions to be drawn here. This is replicated in research on markets around the world, specifically focused on the impact of the high penetration of renewables \citep{Mays2022} \citep{Leslie2020}.  
Overall, electricity price forecasting is an essential part of efficient market operations, and this is only more important in the current and future era of price volatility.
\subsection{Evaluation of Data Used In Literature}
\label{sec:dataused}
There has been a limited amount of research done on forecasting prices on the integrated single electricity market in Ireland, but the research done still contains a variety of different models, data features and time periods used. This is also a field that is growing rapidly, so there is a lot to be learned from looking at international examples. 

What data is used is a critical part of building an accurate model. With the increased penetration of wind and solar power onto the grid, it is now necessary to bring in data to represent them, rather than just focusing on the price of natural gas or oil, or the total demand or generation.
SEMO has a significant amount of data available on its website for free, from forecast wind speeds to contracted wind generation, generation and demand \citep{SEMOpx2023}.
\subsubsection{Wind Energy}
The increased amount of wind energy has been found to decrease the cost of electricity in Ireland \citep{Denny2017}\citep{Higgins2015}. From looking at countries like Germany, it can be seen that off-peak hours (12AM – 9AM and 9PM – 12AM) are most sensitive to the downward price pressure of cheaper wind energy \citep{Rintamäki2017}. The “stochastic character of weather conditions” means that despite wind energy reducing prices on average, it increases the volatility of the electricity markets, making forecasting more difficult \citep{Rintamäki2017}\citep{Acaroğlu2021}. This exasperates Ireland’s existing volatility, which between 2015 – 2019, was one of the highest in Europe \citep{Zakeri2022}.

While the SEMO data can give the contracted wind generation, other research used wind speed forecasts from Met Éireann. The wind speed is a difficult variable to account for, as stated in \citep{Mohamed2022}, using wind speed forecasts can increase the error, as the forecasts themselves contain an intrinsic error. They improve the impact of the forecast by averaging the values of the wind speed at Mullingar, Sherkin Island, Malin Head and Mace Head for their wind speed. This is done to improve the relevance of the wind speed as it considered where most wind farms are located to take advantage of the prevailing south-westerly wind when deciding the weather stations to use \citep{Clarke2022}. Wind speed in Galway, Cork and Dublin has also been found to be highly correlated with the spot price. In \citep{Lynch2021}, it was found that the variable with the highest Pearson Correlation Coefficient (PCC), a measure of linear correlation, was windspeed in Galway, with a value of 0.315, this was followed by the natural gas price at 0.259. For forecasting prices beyond the day-ahead horizon, the correlation between wind and price is strong at the beginning of the forecasting horizon, but this begins to decrease rapidly as the target day is more than four days away\citep{Sgarlato2023}.
 
\subsubsection{Natural Gas Prices}
Natural gas is playing an increasingly important part in the generation mix in Ireland and is set to continue an important role in the future. In 2021, natural gas generated 46 percent of the all-island electricity supply \citep{GNI2022}. In \citep{Lynch2021}, the daily Henry Hub Natural Gas spot price was used in their model and it was found to have a Pearson Correlation Coefficient of 0.259 over the year 2019, not as correlated as windspeed in Galway or Cork, but still clearly impactful. In this same study, daily oil prices only had a PCC of 0.088. 
Research from Turkey has found that natural gas prices have a lot less of an impact on the price of electricity during off-peak hours. This lines up with the research in \citep{Rintamäki2017}, which shows that off-peak hours are when renewables have more of an impact \citep{Poyrazoglu2019}.

\subsubsection{Time-Based Features}
Time-based features explain when the model is forecasting for, whether it’s the year, month, day or hour. In \citep{Arci2018}, the authors consider making different models for each day type, or else making the day type an input variable. The intriguing concept of using separate models for different days was not explored further; instead, the day type was incorporated as an input variable.   
It is also important to consider holiday days like Bank Holidays, but holidays themselves make up only 2.5 percent of the year, so it is recommended to us a long period of training data to accurately account for them \citep{Mohamed2022}. As is also pointed out in \citep{Mohamed2022}, the holidays between Northern Ireland and the Republic of Ireland can differ, leading to additional challenges. Of course, if the electricity price patterns on holidays are similar enough to Sundays that they can be accounted with them, then that could be a stronger dataset, even with a limited period of training data.
\subsection{Evaluation of Training and Testing Periods}
The amount of data given to a model is critical in improving its understanding of the information it has received. This must be balanced with what is said in \citep{Arci2018} that recent data is more useful than old data, and a “very large training set which includes old data is less useful to track the most recent trends”. What that means practically, is that the period of lockdowns of 2020-2021 may not give much useful data for the period when we see the rise in the price of natural gas towards the end of 2021. 
In \citep{Lynch2019} the periods evaluated are 2010-11, 2015-16, and 2016-17. For \citep{Lynch2021} and \citep{O'Leary2021} the models used are trained and validated on 90 percent of the data from 30th September 2018 – 12th of December 2019 and tested on 10 percent of it, with the splits chosen randomly. This is a training and validation set of 394 days and a test set of just 44 days. A similar method is used for \citep{Lynch2021} on the periods 1st January – 12th December 2019, and 1st January - 31st December 2020. [7] is trained over the entirety of 2019 and 2020 and tested on the first three months of 2021.
As can be seen, these in a wide variety of testing periods in research on price forecasting for the I-SEM. However, given the different time periods used, it is difficult to say from this research what length of a training period is appropriate. It is also clear there has not been published research done on day-ahead market price forecasting trained on data from 2021 for the I-SEM, leaving an important period that has to be understood.

\subsection{Evaluation of Models Used In Literature}
There are a wide variety of models available to use for analysing data now, from statistical methods to probabilistic methods, from linear regression to artificial neural networks, there is a wide potential for improving results by improving the choice of models used. Some of the papers written were focused exclusively with the aim of trying different models. So, we have a wide range of models used on the I-SEM.
The study in \citep{Lynch2021} uses a gradient-boosted model, extreme gradient-boosted model and light gradient boosted model and achieved an impressive MAE score of just 9.93 with the extreme gradient-boosted model with 44 days used as a test dataset. In \citep{Lynch2019}, a k-SVM-SVR model (an ensemble model using k-means, SVM and SVR) is used, and achieves impressive results, testing on 2016/2017 data. A variety of models is used in \citep{O'Leary2021}, both machine learning models and deep learning, and uses the same training period and test split as \citep{Lynch2021}. It found that the simpler models outperformed the more complex ones. The models with the lowest MAE score were K-Nearest Neighbours Regression (KNR), Linear Regression and Random Forest, with MAE scores of 11.21, 11.52 and 11.54, respectively. A lot of different models have been used, although the different time periods analysed means they are difficult to directly compare.
Most of the papers do not go into detail about the parameters of the models used, which makes it more difficult to reproduce the findings. This is an ongoing issue in electricity price forecasting research \citep{Lago2021}. 
\subsection{Objectives and Structure of This Study}
\label{sec:overview_researchobjectives}
The primary aim of this paper is to examine how optimal forecasting methodologies for the day-ahead market within the I-SEM have changed in light of the escalating volatility of prices. This study specifically focuses on the evolution of different forecasting models and the extent of training data employed, aiming to ascertain whether the most effective techniques from early 2020 maintain their accuracy in late 2022, or if different models should be used. To the authors knowledge, research has not been published for the I-SEM that tries different lengths of training periods to see what is the most accurate. 

This study bears significant implications for the comprehension of the most appropriate methodologies during periods of price fluctuations in the present and future, as is predicted will be necessary. By identifying effective forecasting strategies for these situations, the study aims to improve efficiency of both electricity retailers and suppliers, ultimately contributing to the mitigation of price increases for consumers.

\section{Data and Methodology}
The key to ensuring the model is accurate is collecting all the important data used to determine the price of electricity. The research in Section \ref{sec:dataused} helped determine what data is useful and what is not. It is difficult to track down the required information from relevant agencies, and then to put it all together, especially for the period spanning the four years from 1st October 2018 – 30th September 2022 \citep{SEMODoc}.
\subsection{Data Sources and Description}
\subsubsection{System Data}
Eirgrid (the system operator of the transmission system), publish files containing lots of useful information, like Generation, Demand, Wind Availability and Generation (all split for Republic of Ireland and Northern Ireland), Solar Availability and Generation and SNSP values \citep{EirgridLib}. System Non-Synchronous Penetration (SNSP) is a measure of the non-synchronous generation on the system at an instant in time, it is calculated based on the volume of non-synchronous energy generated plus interconnector imports as a percentage of the overall demand plus interconnector exports \citep{Gaffney2019}. These values are in quarter-hourly values, the average of these for the hourly periods were used, and combined with the relevant time period from the price data from SEMOpx \citep{SEMOpx2023}. This was done for the period of 1st Oct 2018 – 30th September 2022.
For the price of Natural Gas, the EU Daily Natural Gas price \citep{TE2023} was used, which is the Dutch Tile Transfer Facility (TTF) price, which serves as a proxy for the EU-wide price \citep{Liboreiro2022}. For days where markets were not open, the market closing value was used.
\subsubsection{Weather Data}
For the data around weather, like air temperature and wind speed, this was comparatively simple to find. Met Éireann’s publish hourly, daily and monthly data for any weather station in the Republic of Ireland \citep{MetEir}.
Wind speed data was collected following the example of \citep{Mohamed2022}, and an average of these values was made. The air temperature in Malin Head and Dublin Airport, and the hours of sunshine from Dublin Airport, were also downloaded.
The data taken from Eirgrid have quarter-hourly data on Northern Ireland Solar Availability and Generation, so there is practical Northern Ireland-specific weather-dependent data used. 
\subsubsection{Time and Date Data}
All of the time data was taken from the SEMOpx dataset. While most of the time features, like day of the month, month of the year, and year, were included in the system data, there was more data to add. Day of the year, day of the week, and week of the year were added as features.
Each year has two non-24 hours days, from clocks going forward an hour in the Spring, and then going back an hour in the Autumn. This means one day each year has only 23 hours, and one day has 25 hours. To avoid overfitting on the 25 hours, as set out in \citep{Pavićević2022}, the 25th hours of those days were removed, and the 23rd hour was repeated, to give each day 24 hours. Another potential way would have been to average the values of the 24th and 25th hours, as set out in \citep{Ziel2018}, but the improvement of this would be minimal.
\subsection{Data Analysis}
After the data was compiled, a sample of features were tested for their PCC values compared to the DAM price over the years 2019 –  30th Sept 2022, to see how the impact of values has changed over the period, this can be seen in Table \ref{tab:corrtable}. 
\begin{table}[hbt!]
  \captionsetup{labelsep=newline,justification=justified,singlelinecheck=false}
  \caption{PCC Values for a sample of features, 2019 - 2022.}
  
  {\footnotesize
  \begin{tabular}{lcccc}
    \toprule
     & \multicolumn{4}{c}{PCC Values} \\
    \toprule
    Feature & 2019 & 2020 & 2021 & 2022 \\
    \midrule
    Total Demand	&0.56&	0.60&	0.39&	0.26\\
    Total Generation	&0.17	&0.31&	0.28	&0.09\\
    Total Wind Generation	&-0.33&	-0.22&	-0.20&	-0.45 \\
    EU Natural Gas Price	&0.33&	0.41	&0.78	&0.69\\
    Wind Speed – Galway	&-0.33&	-0.27&	-0.12	&-0.44\\
    NI Solar Generation	&0.11	&-0.02	&-0.01&	0.14\\
    SNSP	&-0.34&	-0.32	&-0.33	&-0.52\\
    Sunshine Hours – Dublin Airport	&0.07&	-0.01	&0.01&	0.09\\
    \bottomrule
  \end{tabular}
  }
  \label{tab:corrtable}
\end{table}
The first thing to note is that the EU Natural Gas, for 2019, has a PCC value of 0.334 compared to the day-ahead market price. Other research on the I-SEM has found that the daily Henry Hub Natural Gas Price has a PCC value of 0.259 for the same period \citep{Lynch2021}. This finding suggests that the daily EU Natural Gas price is a more useful feature for electricity price forecasting in Ireland than the daily Henry Hub Natural Gas price. 
It can be seen that SNSP is highly correlated with the day-ahead market price, and that renewables are pushing down the price of electricity. SNSP may be more correlated as it reflects the actual amount of renewable generation used, not just the total amount generated. Eirgrid maintains a maximum of 75 percent, to ensure grid stability \citep{Eirgrid2022}, so any generation past this cannot be used. This is another reason why SNSP is a more correlated value.

Another noticeable finding is that solar generation (data is only available for Northern Ireland), is not always negatively correlated with price. This could be due to the fact that solar generation output is highest during the day, when the price is higher, and the solar generation is not having enough of an impact to push down the price, unlike wind energy. Both solar generation and hour of sunshine have very low levels of correlation. 

\begin{figure}[h ]
    \centering
    \includegraphics[width=1\linewidth]{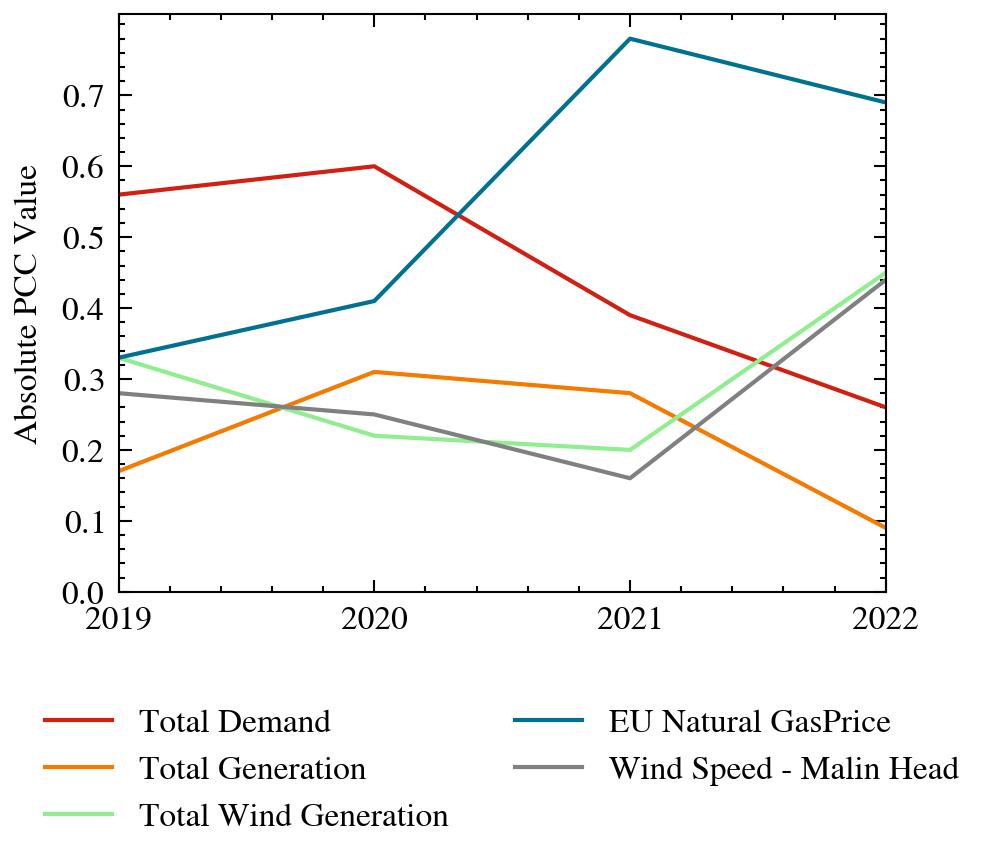}
    \caption{Absolute PCC Values for features compared to DAM price, 2019-2022}
    \label{fig:pcc}
\end{figure}
Looking at Figure \ref{fig:pcc}, we can see there are some notable trends in this data. Firstly, while demand was the feature with the highest PCC value, a lot higher than the PCC value for generation, both of those have fallen in recent years. In comparison, the PCC values for natural gas price, wind generation and wind speed have risen over the same period. This tells us that the input fuel for electricity has become a more important driver of the price of it, than the total generation or demand, however, demand is still a more correlated value than generation. 
\subsection{Feature Selection}
Firstly, a Linear Regression model was trained with all of the features collected, on the years 2019, 2020 and 2021, and tested over 2022 (Jan – Sept). All the data was scaled using scikit-learn’s StandardScaler function \citep{sciscaler}, before the model was ran.
After the model was run, the coefficients of the features were analysed, and the features with the lowest coefficient were iteratively removed until the most accurate set of parameters was found. This method of backward elimination is a commonly utilised methods of feature selection for electricity price forecasting \citep{Lago2021}\citep{Uniejewski2016}. This resulted in the removal of hour, day of week, day of month, Sunshine Hours in Dublin Airport, Air Temperature (Dublin Airport and Malin Head), Average Wind Speed (Total and Mullingar) and Total Demand 2 Days ago. The full list of features used can be seen in Table \ref{tab:featuretable}.
\begin{table}
\captionsetup{labelsep=newline,justification=justified,singlelinecheck=false}
  \caption{All features used in the finals models.}
  
  {\scriptsize
  \begin{tabular}{lll}
    
    \toprule
     \multicolumn{3}{c}{Forecasting Model Features} \\
 \multicolumn{3}{c}{\textbf{IE:} Republic of Ireland                         \textbf{NI:} Northern Ireland}\\
    \midrule
    Week	&  Day of the Year&	DAMPrice 1 Day Before\\
    EU Natural Gas Price&  IE Demand&IE Wind Availability\\
    SNSP&  NI Demand&DAMPrice 2 Days Before\\
    Total Demand&  Total Demand 1 Day Before&DAMPrice 1 Week Before\\
    NI Wind Availability&NI Generation&IE Wind Generation\\
    Total Wind Generation&Month&Total Generation\\
    Demand 1Week Before&NI Wind Generation&IE Generation\\
    NI Solar Generation&Year&\\
    &\textbf{Average Wind Speed:}&\\
    Dublin Airport& Mace Head& Malin Head\\
    \bottomrule
  \end{tabular}
  }
  \label{tab:featuretable}
\end{table}

The model trained on all the parameters has a mean absolute error of 48.52 while this was reduced to 47.92 by the smaller group of parameters, a reduction of 1.24 percent. It can be understood from this that more data is not always better, and that some data features are not to be used when forecasting.
\subsection{Methodology for Testing and Evaluating Model Performance}
To analyse the models, the various models are tested over 11 different periods: January – March, April – June, July – September and October - December for 2020, 2021 and 2022 (with the exception of no October – December period for 2022 as the data considered in this paper only extends to Sept 30th 2022). It is well-established in electricity price forecasting that periods of a week or even weeks are too short to properly analyse a model \citep{Lago2021} \citep{Aggarwal2009} and as stated in \citep{Weron2014}, “Only longer test samples of several months to over a year should be considered.” Each model will be tested on all 11 of these periods and reviewed using the metrics outlined in Section \ref{sec:metrics}. 

Furthermore, each model is trained on 3 different training periods, so that it can be understood if more data is better for accuracy. This will consist of six months before, one year before, and two years before (for the 2020 models this will be slightly less than two years due to the absence of data). 

This will give a total of 33 different periods for each of the models that are analysed, showing a clear picture of how models perform with different training and test periods. This will also explain if the models that were the most accurate with the test periods in 2020 are the most accurate in recent years. The training periods created for each testing period can be seen in Table 4.
\begin{table}[hbt!]
  \captionsetup{labelsep=newline,justification=justified,singlelinecheck=false}
  \caption{Training periods used for each testing period.}
  {\footnotesize
  \begin{tabular}{cccc}
    \toprule
    Test Period&Train Period 1&Train Period 2	&Train Period 3\\
    \midrule
    2020 Jan - Mar	&6 Months &	1 Year&	14 Months\\
    2020 Apr - Jun&6 Months &1 Year&	17 Months\\
    2020 Jul - Sep	&6 Months&1 Year&20 Months\\
    2020 Oct - Dec&	6 Months&	1 Year&2 Years \\
     
\begin{tabular}{@{}c@{}}
    2021 \& 2022 \\
    – All Quarters
    \end{tabular}
    &6 Months&1 Year&	2 Years \\
    \bottomrule
  \end{tabular}
  }
  \label{tab:testperiod}

\end{table}

\section{Models and Forecasting Techniques}
Electricity price forecasting is a field that uses many types of computational models for price prediction, ranging from simpler regression models to complex deep learning models. Despite the increased use of these more complex models, the research in Ireland has found that they are not more accurate for price forecasting \citep{O'Leary2021}. This section aims to explain some of the basis of these methods, before they are implemented later.
\subsection{Overview of Machine Learning and Neural Network Models}

\subsubsection{Linear Regression}
Linear Regression is a widely used statistical method for predicting a continuous output variable from one or more input variables. The linear regression model assumes that there is a linear relationship between the input variables and the output variable, and seeks to find the best-fitting line through the data points. The line is defined by the intercept and the slope coefficients, which are estimated from the training data.
Linear regression was the first method attempted for electricity price forecasting \citep{Jedrzejewski2022}, and is still an important part of the forecasting selection of researchers and market participants.
\subsubsection{Random Forest Regression}
Random Forest is a machine learning algorithm that is commonly used for classification tasks, but it can also be applied to regression problems. It works by creating a collection of decision trees, where each tree is trained on a random subset of the features and a random subset of the training data. The output is generated by taking the average of the predictions from all the decision trees in the forest \citep{Pórtoles2018}.
The performance of a random forest model can be adjusted by changing various parameters, such as the number of decision trees, the maximum depth of the trees, and the maximum number of features that can be considered per tree \citep{Dudek2022}.

\subsubsection{Gradient Boosted Machines}
Gradient Boosting is an ensemble machine learning method that uses decision trees as its foundation, sequentially adding trees to iteratively minimise the errors. By combining the output of multiple ‘weaker’ learners, gradient boosting builds a more accurate, ‘stronger’ model \citep{Touzani2018}. Extreme Gradient Boosting, first proposed in 2016 \citep{Chen2016}, is a modified method that has worked very well for price forecasting in the I-SEM \citep{Lynch2021} and in a widespread range of prediction tasks. This is an optimised and regularised version of the original version, helping to control overfitting \citep{Carmona2019}. It also uses simplified objective functions that enables the integration of the predictive and regularisation terms while optimising computational resources \citep{Fan2018}.

Overall, Gradient Boosting, and especially Extreme Gradient Boosting, are powerful and versatile methods that are popular choices with electricity price forecasting and other forecasting tasks.
\subsubsection{Support Vector Machines}
Support Vector Machines is a versatile model that can be used for linear and non-linear approaches to classification or regression. A well-established method, it had a breakthrough in the early 1990’s \citep{Boser1992}, and has widely used since. When it is used for regression (known as SVR), its objective is to construct a hyperplane that fits as close to the data points as possible in a way that allows for a specified error tolerance \citep{Trafalis2000}, striking a balance between accurately capturing the trends in the data and avoiding overfitting. As a regression tool, it has the advantage of finding the global optimum where other methods may get stuck in local minima. This is due to its formulation as a convex quadratic optimisation problem \citep{Sánchez2011}.

The model maps the input data into a higher-dimensional feature space through the use of a kernel function. This transformation allows for the identification of non-linear relationships between the input features and the target variable. It can be a linear or non-linear regression method depending on the kernel function used \citep{Boucher2015}.

Linear SVR and SVR have been shown to be similarly capable at price forecasting on the I-SEM \citep{O'Leary2021}.
\subsubsection{K-Nearest Neighbours}
K-Nearest Neighbours is well-known as a model for unsupervised classification, but it can also be used for regression (known sometimes as KNR or KNN Regression). It works on the assumption that data points in close proximity to each other are likely to have a similar output. When given a set of data we want the price for, it works by aggregating the target values of its k-nearest neighbours to predict the unknown target value \citep{Martínez2018}. Calculating the optimum values for predictor weights and number of neighbours is important to building a useful model \citep{Modaresi2018}. Too many neighbours and the model may not capture important patterns, too few neighbours may lead to overfitting.

\subsubsection{Perceptron \& Multilayer Perceptron}
First implemented in 1958 \citep{Lefkowitz2019}, a perceptron is a model of a biological neuron. It's relatively simple strucutre can be seen in Figure 6. It is a type of artificial neuron that has numbers for inputs and outputs, rather than binary values \citep{Géron1}. Each input connection has a weight that is changed to improve the accuracy of the model. The formula inside the model can be changed depending on the task at hand. It is normally used for binary classification, but with the activation function being set to a linear activation function, it can be used for linear regression.
\begin{figure}
    \centering
    \includegraphics[width=1\linewidth]{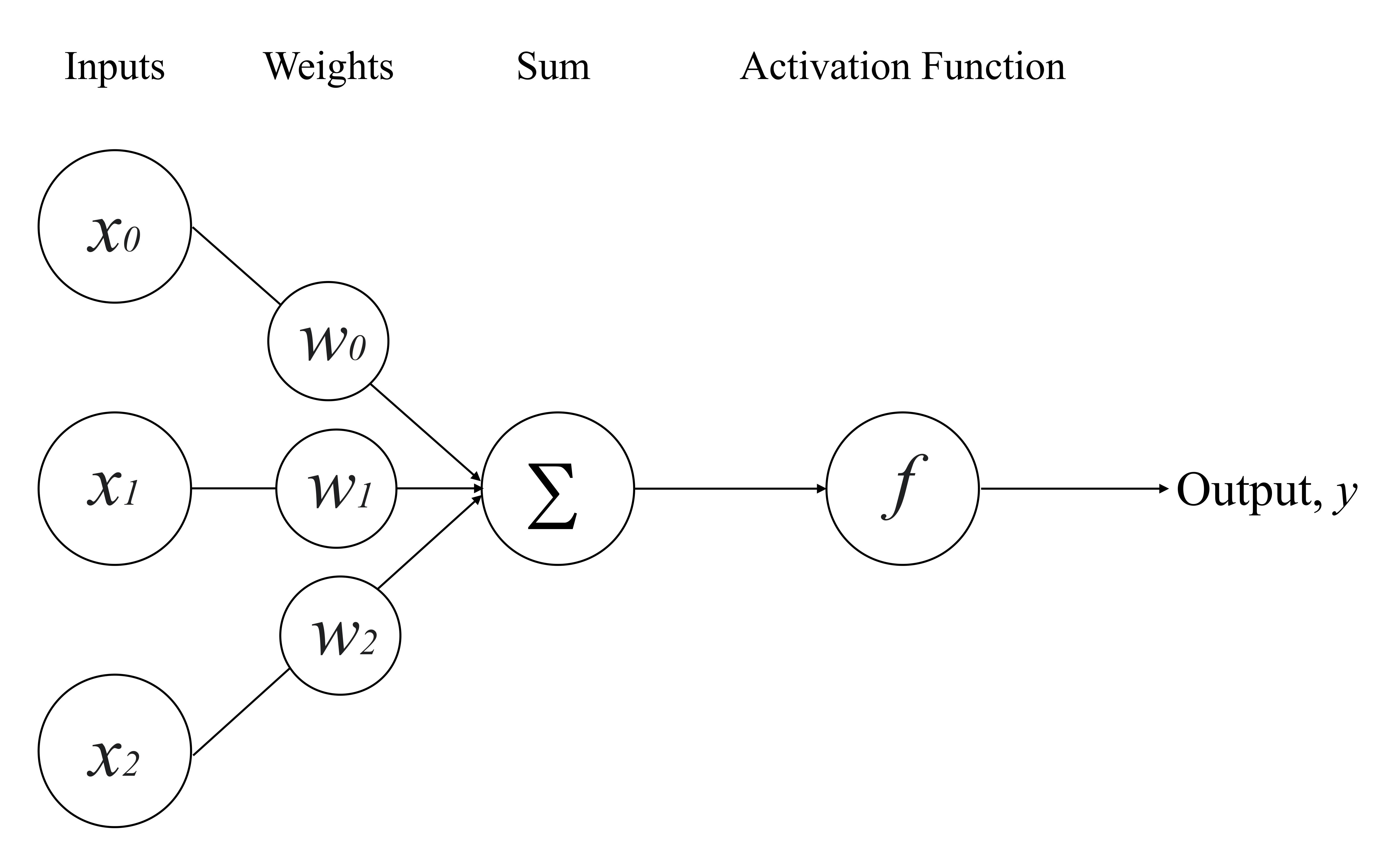}
    \caption{Model of a Perceptron with multiple inputs}
    \label{fig:perceptron}
\end{figure}
\begin{figure}
    \centering
    \includegraphics[width=1\linewidth]{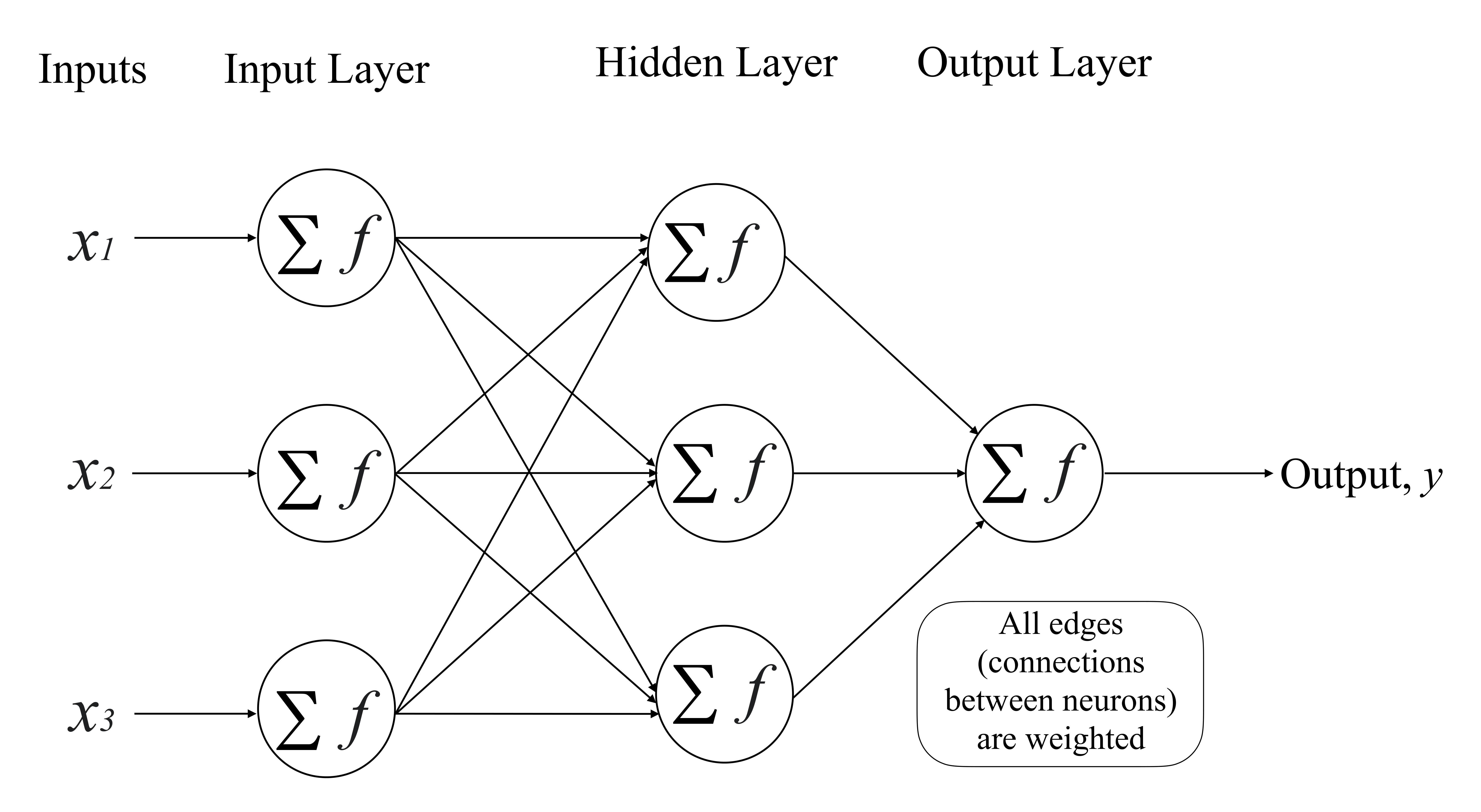}
    \caption{Model of a Multi-Layer Perceptron}
    \label{fig:mlp}
\end{figure}
As seen in Figure 7, a Multilayer Perceptron is when these perceptrons are constructed into multiple layers, composing an input layer, an output layer, and at least one hidden layer. Each layer contains neurons with activation functions, which transform the weighted sum of their inputs before passing the result to the next layer. The connections between neurons have weights that are updated during training. MLPs are a specific type of ANN, but there are also many different types of neural networks, that will be explored in the rest of the section.

\subsubsection{Long-Short Term Memory}
Long Short-Term Memory neural networks are a type of recurrent neural network (RNN) architecture specifically designed to overcome the challenges associated with learning long-term trends by RNNs, known as the vanishing gradients problem. LSTMs have proven to be highly effective in addressing various forecasting tasks, due to its ability to recognise impactful inputs and preserve it for as long as needed, making it useful for long-term pattern recognition in time series \citep{Géron2}.

The fundamental building block of an LSTM network is the LSTM cell, which incorporates a series of gating mechanisms to control the flow of information throughout the network. These gates include the input gate, forget gate, and output gate, each playing a distinct role in  determining the extent to which new information is incorporated into the cell, deciding the degree to which the previous cell state is retained or discarded, and controlling the contribution of the cell state to the hidden state.

The gating mechanisms inherent to LSTMs allow them to selectively store, update, or discard information, thereby mitigating the vanishing and exploding gradient problems that commonly plague traditional RNNs.

In the context of electricity price forecasting, LSTMs can effectively capture temporal dependencies and model non-linear relationships within the data \citep{Kawabata2020}. Given the inherently time-sensitive nature of electricity prices, influenced by factors such as demand, generation and weather conditions, LSTMs are well-suited for this task.

\subsubsection{Gated Recurrent Units}
This is essentially a simplified version of the LSTM \citep{Géron2}. A GRU only has an update gate and a reset gate, these are used to control how much new information is incorporated, and how much of the previous hidden state is exposed to the next hidden state. This decrease in resources makes it more computationally efficient than LSTM, and performs just as well for electricity price forecasting \citep{O'Leary2021} and performs comparably, if not better on a host of other prediction tasks as well \citep{Chung2014}. 

\subsubsection{Convolutional Neural Network }
A Convolution Neural Network, seen in Figure 8, is a multilayer network. Like LSTM, it was created to  address the vanishing gradient problem in RNNs. It can have three different types of layers: a convolutional layer, a pooling layer, and a fully connected layer \citep{IBM1}. The convolutional layer serves to capture local patterns in the data, the pooling layer summarises the extracted features (mostly by giving the maximum value for sub-region \citep{Albawi2017}), with these then being flattened to a vector for the fully connected layer. These serve to distinguish the important parts of the input data, extracting the important features, from the large amounts of data. 
\begin{figure}[hbt!]
    \centering
    \includegraphics[width=1\linewidth]{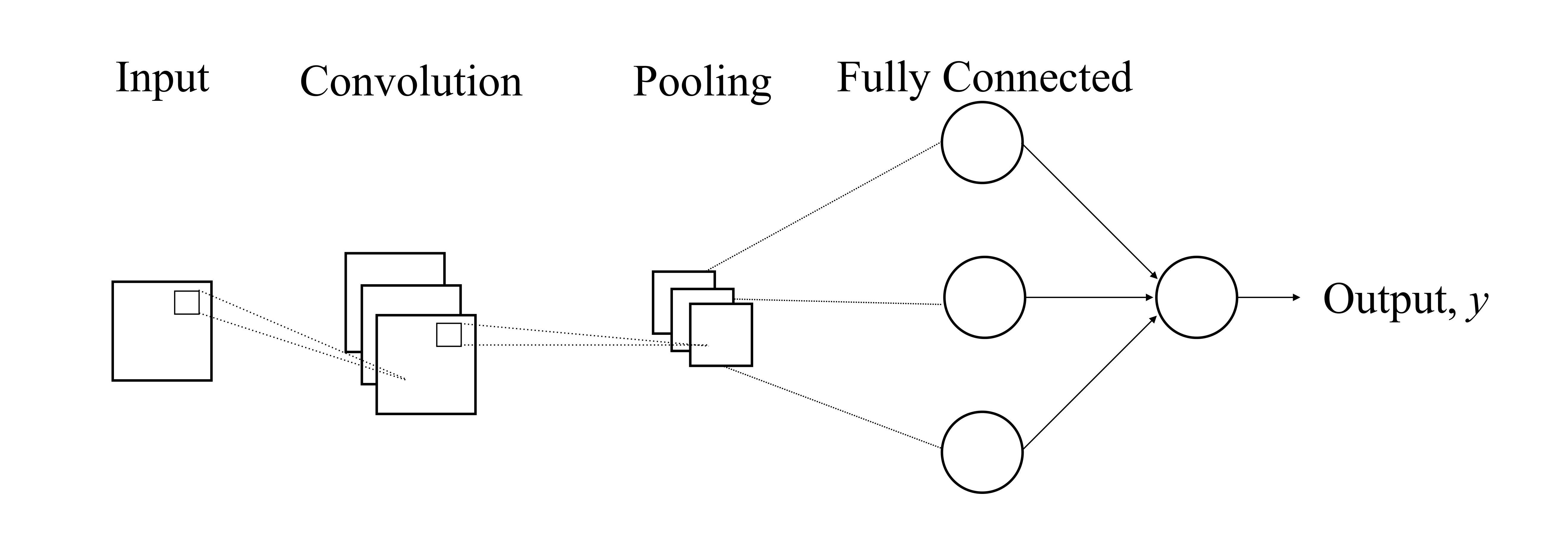}
    \caption{Model of a Convolutional Neural Network}
    \label{fig:cnn}
\end{figure}

While CNNs have being used mostly for image recognition, they have been used successfully for electricity price forecasting in Ireland \citep{O'Leary2021} and elsewhere \citep{Khan2020}. 

\subsection{Comparative Study of Different Model Architectures}
The perceptron, convolutional neural network, gated recurrent unit and long-short term unit are all taken as a sample of the most accurate neural network models from \citep{O'Leary2021}, they have the same structures as are outlined in the paper. The ANN with one hidden layer is taken from the day-ahead market price forecasting section of \citep{Mohamed2022}, and is compared with an ANN with three hidden layers. When parameters aren't made clear in the original papers, an iterative process is used to select the most accurate parameters for each model, using the same training and test set that was used to select the features.
This is a comprehensive overview of the models used for electricity price forecasting in Ireland and in other countries. The full list of the models used can be found in Table 5. Analysing all of these should explain whether traditional Machine Learning models or more complex Neural Network models work better in recent, more volatile times.
\begin{table*}[hbt!]
  \captionsetup{labelsep=newline,justification=justified,singlelinecheck=false}
  \caption{Parameters used for the ML and NN models used  for experiments.}
  \label{tab:paramtable}
  {\footnotesize
  \begin{tabularx}{\textwidth}{XXll}
    \toprule
    Name& Model Type&Parameters& Used In\\
    \midrule

	Linear Regression& Linear Regression& &	\citep{O'Leary2021}\\
    \midrule
    Random Forest & 
    Random Forest &
    \begin{tabular}[t]{@{}l@{}}n\_estimators = 70; \\max\_depth = 8;\\min\_samples\_split = 1; \end{tabular} &	
    \citep{O'Leary2021}\\
    \midrule
    xGBRegressor & 
    Extreme Gradient Boosted Regressor &
    \begin{tabular}[t]{@{}l@{}}learning\_rate = 0.05; \\ max\_depth = 5;\\n\_estimators=100; \end{tabular} &	
    \citep{Lynch2021}\\
    \midrule
    LinearSVM & 
    Linear SVM &
    \begin{tabular}[t]{@{}l@{}}epsilon = 0.07; \\loss = 'squared\_epsilon\_insensitive';\\C = 1.5;\\max\_iter=7500;\end{tabular} &	
    \citep{Lynch2019}\citep{O'Leary2021}\\
    \midrule
    KNNRegressor & 
    KNNRegressor &
    \begin{tabular}[t]{@{}l@{}}n\_neighbors=11; \\weights = 'distance';\\C = 1.5;\\algorithm= 'auto';\\leaf\_size=5;\end{tabular} &	
    \citep{O'Leary2021}\\
    \midrule
    dense0 &
    Perceptron& 
    \begin{tabular}[t]{@{}l@{}}Single Perceptron; \\ Linear Activation Function;\end{tabular}
     &	
    \citep{O'Leary2021}\\
    \midrule
    gru1 &
    Gated Recurrent Unit & 
    GRU unit (16 neurons) &	
    \citep{O'Leary2021}\\
    \midrule
    lstm1 &
    Long-Short Term Memory Unit& 
    LSTM unit (16 neurons)&	
    \citep{O'Leary2021}\\
    \midrule
    cnn4&
    Convolutional Neural Network & 
    \begin{tabular}[t]{@{}l@{}}
    1D convolutional (64 filters),\\
    1D convolutional (32 filters),\\
    Max pooling,\\
    flatten,\\
    dense (16 neurons),\\
    output neuron
    \end{tabular} &	
    \citep{O'Leary2021}\\
    \midrule
    4n\_hidden &
    Multi-Layer Perceptron & 
    \begin{tabular}[t]{@{}l@{}}
    Four Neurons,\\
    Output Neuron
    \end{tabular} &	
    \citep{O'Leary2021}\\
    \midrule
    multiple\_hidden &
    Multi-Layer Perceptron& 
    \begin{tabular}[t]{@{}l@{}}
    32 neurons,\\
    64 neurons,\\
    32 neurons,\\
    output neuron
    \end{tabular} &	
    \\
    \bottomrule  
  \end{tabularx}
  }
  
\end{table*}

\begin{figure}[hbt!]
    \centering
    \includegraphics[width=1\linewidth]{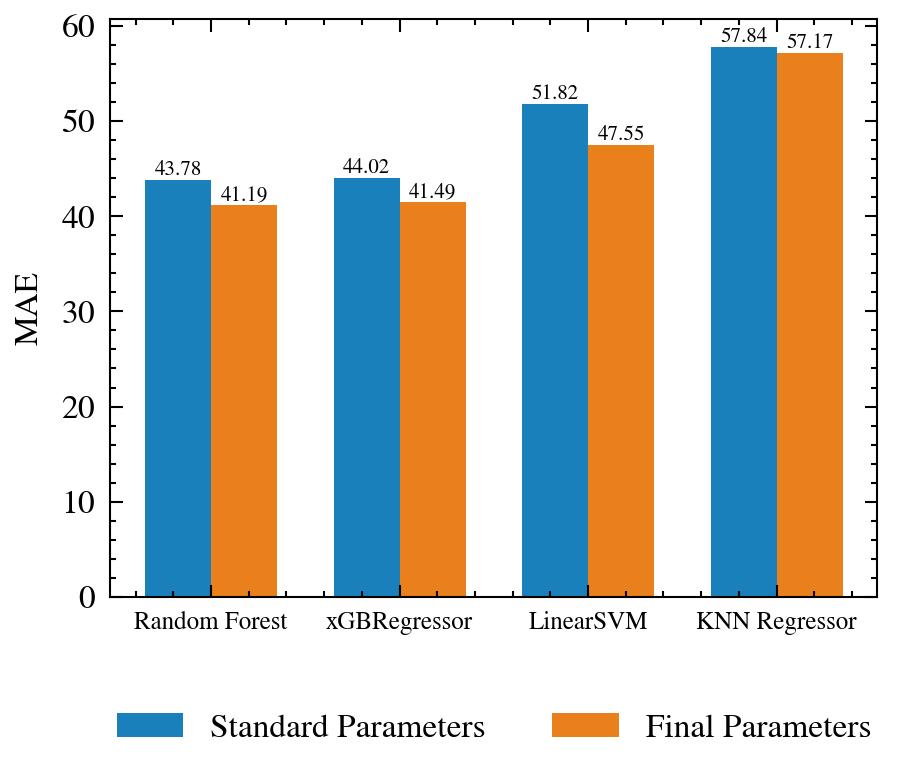}
    \caption{Change in MAE for the ML models after parameter optimisation}
    \label{fig:paramopt}
\end{figure}

The parameters chosen for each model is done in a similar way to how the features used were selected. The range of options for each of the main parameters for each model was created, and the model iteratively ran through these options, with the lowest MAE indicating the optimal parameters. Although this took some time, it did bring improvements in the models.
Not all models improved by as much however, the xGBRegressor improved by 5.75 percent, while the KNNRegressor only improved by 1.17 percent. This difference in improvement can be seen in Figure \ref{fig:paramopt}. 

The machine learning models were all implemented using their respective scikit-learn implementations, and TensorFlow \citep{Abadi2016} was used for the neural network models.

\section{Results and Discussion}
\subsection{Evaluation Metrics}
\label{sec:metrics}
Each evaluation of electricity price forecasting models is essential to understanding their performance and reliability. In this section, the different metrics used to assess the models will be discussed, methods like Mean Absolute Error, Root Mean Squared Error, and Relative Mean Absolute Error. They are all negatively orientated metrics, which means the lower the metric result the better, with 0 indicating a perfect fit. Insights into the strengths and weaknesses of these metrics will be provided, which will aid us in the selection of the most accurate model. 
Mean Absolute Percentage Error is not used in this research as there are a number of issues with it, including dealing poorly with price spikes \citep{Weron2014}, and with values close to zero \citep{Boylan2021}, both of which are frequent occurrences in electricity markets. It also favours methods which under-forecast, a problem with the era of more regular price spikes \citep{Tofallis2013}.
\subsubsection{Mean Absolute Error}
Mean Absolute Error (MAE) is a standard error metric in electricity price forecasting and in forecasting tasks across all industries. It is defined as the average of the absolute difference between real and predicted values. The value for MAE is in the same scale as the dataset, and has the advantage of being intuitive to understand. However, since it is an absolute value, it is difficult to use it to compare predictions on different datasets with \citep{Weron2014}\citep{Lago2021}. 

\begin{equation}
\text{MAE} = \frac{1}{n} \sum_{i=1}^{n} \left| y_i - \hat{y}_i \right|
\end{equation}
\subsubsection{Root Mean Squared Error}
Root Mean Squared Error (RSME) is another metric that is a standard in the field of electricity price forecasting. It is calculated as the square root of the average squared error between the real value and the predicted output. It does not deal with large outliers well, as it squares the difference, whereas MAE uses the absolute value. It does have the same problem as MAE, in that it is difficult to be used to compare forecasts of different datasets, as it is not scaled \citep{Lago2021}.

\begin{equation}
\text{RMSE} = \sqrt{\frac{1}{n} \sum_{i=1}^{n} (y_i - \hat{y}_i)^2}
\end{equation}
\subsubsection{Relative Mean Absolute Error}
While different versions of this metric are used in other areas of research \citep{Hyndman2006}, this was first used for electricity price forcasting in \citep{Lago2021}. To the authors knowledge, it has never been used on price forecasting for the I-SEM before. It normalises the MAE by the MAE of a naïve forecast (estimating technique where the value for the future period is set at the value of the current period). The naïve forecast was set at a daily level, so the day-ahead market price for 11pm would be equal to the value of 11pm the previous day, and this continues for every hour. This metric allows us to better understand how models perform across time periods with different ranges of prices, which can be seen on the huge increase in prices across recent years. This metric will make clear whether models are getting more inaccurate in recent periods, or if they rise in MAE and RMSE is just due to the rise in prices.
\begin{equation}
\text{rMAE$_m$} = \frac{1}{N} \sum_{k=1}^{N} \frac{|p_k - \hat{p}_k|}{\frac{1}{N-m} \sum_{i=m+1}^{N} |p_i - p_{i-m}|}\\
\end{equation}
where $m$ represents the seasonal length (in the case of day-ahead prices that could be either 24 or 168 representing the daily and weekly seasonalities).

This is built in Python using the epftoolbox, using the rMAE function \citep{Lago2020}. For the parameters of the function, seasonality is set to daily (‘D’) and frequency is set to hourly (‘1H’).

\subsection{Analysis of Forecasting Accuracy Across Different Models}
After all models had been tested 30 times, the averages of the error metrics were taken. These results can be seen in Table 6. The primary outcome observed from the analysis is an increased difficulty in forecasting electricity prices. This trend is evident across various methods and training periods, as demonstrated by a consistent rise in all performance metrics over the 11 quarters examined. 
Furthermore, the fluctuation in these metrics indicates a shift in the accuracy of forecasting models and optimal training periods over the span of three years.
\begin{table}[h]
  \captionsetup{labelsep=newline,justification=justified,singlelinecheck=false}
  \caption{The best performing model and training period for each test period.}
  {\scriptsize
  \begin{tabular}{llcccc}
    \toprule
    Test Set&Model	&Training Period&MAE&	RMSE	&rMAE\\
    \midrule
    2020 Q1	&xGBRegressor	&3&	7.50	&10.56	&0.61 \\
    2020 Q2	&xGBRegressor	&1	&5.33	&6.97	&0.61\\
    2020 Q3	&xGBRegressor	&3&	5.69	&9.07	&0.54\\
    2020 Q4	&Random Forest 	&3&10.76	&18.63	&0.61\\
    2021 Q1	&xGBRegressor	&3	&11.96&	23.75&	0.65\\
    2021 Q2	&dense0&3&	12.80&	22.48&	0.66\\
    2021 Q3	&L. Regression \& L.SVM&	2 &	17.76&32.32&	0.75\\
    2021 Q4	&dense0	&2&	28.28	&38.79&	0.65\\
    2022 Q1	&dense0	&2&35.13	&54.81&	0.62\\
    2022 Q2	&xGBRegressor	&2 &37.72&	47.85&	0.95\\
    2022 Q3	&dense0	&1&	37.15	&46.92	&0.79\\
    \bottomrule
  \end{tabular}
  }
  \label{tab:testperiod1}
\end{table}
In even the best-performing models, there is a notable near five-fold increase in the Mean Absolute Error (MAE) and a similar escalation in the Root Mean Square Error (RMSE). However, this substantial increase is not mirrored in the relative Mean Absolute Error (rMAE), suggesting that while the models' accuracy has diminished over time, they are not as imprecise as the MAE alone might imply. This discrepancy underscores the significance of scaled metrics and highlights the practical utility of rMAE in evaluating model performance.
From the second quarter of 2021 on, the predominant model is dense0, the single perceptron with a linear activation function. This is the most fundamental of artificial neurons. More complex models are not providing more accuracy.
An interesting note is that the LinearSVM model performed exactly as well as Linear Regression, with the same errors for all predictions. It can be assumed that the optimised parameters selected caused it to behave like the Linear Regression  model. 
It can also be seen that more data is not always providing more accuracy either. From the second quarter of 2021 on, using the most available data only provided the most accuracy in a 33 percent of the cases. Compared to the quarters before this, using all the available data made the most accurate model in 60 percent of the periods.
Looking at the five best performing models for the third quarter of 2022, in Table \ref{tab:testperiod2}, shows the predominance of the linear models. Table \ref{tab:testperiod3} shows the worst performing models for this period. 
\begin{table}[h]
  \captionsetup{labelsep=newline,justification=justified,singlelinecheck=false}
  \caption{Top five performing models and training periods for 2022 Q3.}
  \label{tab:testperiod2}
  {\scriptsize
  \begin{tabular}{llcccc}
    \toprule
    Test Set&Model	&Training Period	&MAE&	RMSE	&rMAE\\
    \midrule
    2022 Q3&dense0	&1&	37.15&46.92	&0.79\\
    2022 Q3&Linear Regression	&1&	37.20&46.79&0.79\\
    2022 Q3&KNNRegressor	&2&44.97&56.37&0.96\\
    2022 Q3&dense0	&2&	47.45&59.14&1.01\\
    2022 Q3&L.Regression \& L.SVM	&2&	47.64&59.23&1.01\\
    \bottomrule
  \end{tabular}
  }
  
\end{table}
\begin{table}[h]
  \captionsetup{labelsep=newline,justification=justified,singlelinecheck=false}
  \caption{Worst five performing models and training periods for 2022 Q3.}
  \label{tab:testperiod3}
  {\scriptsize
  \begin{tabular}{llcccc}
    \toprule
    Test Set&Model	&Training Period	&MAE&	RMSE	&rMAE\\
    \midrule
    2022 Q3&cnn4&3&159.38&182.10&3.39\\
    2022 Q3&cnn4&2&120.23&150.69&2.56\\
    2022 Q3&cnn4&1&117.14&148.28&2.49\\
    2022 Q3&Random Forest &1&112.70&130.99&2.39\\
    2022 Q3&lstm1&3&111.34&136.66&2.37\\
    \bottomrule
  \end{tabular}
  }
  
\end{table}

\begin{figure}[hbt!]
    \centering
    \includegraphics[width=1\linewidth]{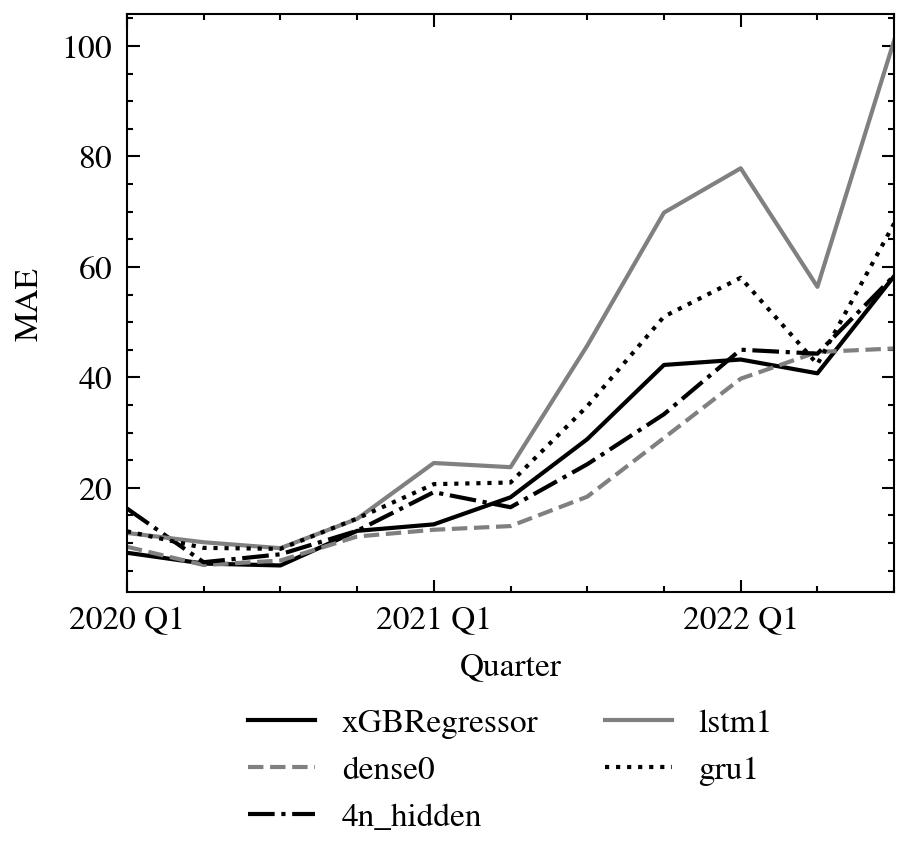}
    \caption{Change in MAE values of a sample models over the 11 test quarters}
    \label{fig:maechange}
\end{figure}

\begin{figure}[hbt!]
    \centering
    \includegraphics[width=0.9\linewidth]{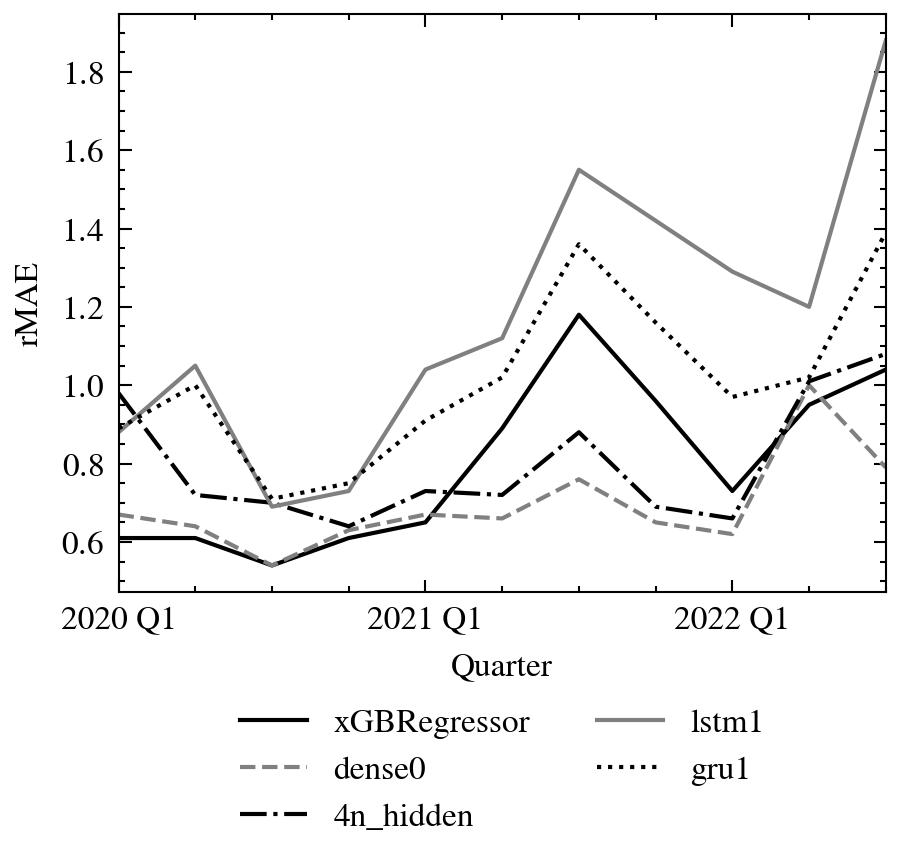}
    \caption{Change in rMAE values of a sample models over the 11 test quarters}
    \label{fig:rmaechange}
\end{figure}

Looking at Table \ref{tab:testperiod4}, with the results for the last quarter of 2021, an extremely tumultuous era, the accuracy of the linear models is clear, as well as the lack of inclusion of Training Period 3.
{
\begin{table}[h]
  \captionsetup{labelsep=newline,justification=justified,singlelinecheck=false}
  \caption{Top five performing models and training periods for 2021 Q4.}
  \label{tab:testperiod4}
  {\scriptsize
  \begin{tabular}{llc c c c}
    \toprule
    Test Set&	Model	&Training Period&MAE&	RMSE	&rMAE\\
    \midrule
    2021 Q4&dense0&2&28.28	&38.79	&0.65\\
    2021 Q4&dense0&1&28.68	&38.11	&0.66\\
    2021 Q4&L. Regression \& L.SVM&2&28.71&	39.06&0.66\\
    2021 Q4&L. Regression \& L.SVM&1&28.89&38.12&0.66\\
    2021 Q4&4n\_hidden&2&29.91&39.83&0.69\\
    \bottomrule
  \end{tabular}
  }
\end{table}
}
Looking at Figure \ref{fig:maechange}, we can see how different models have changed over time. This graph shows the best MAE value for each quarter for a range of models. Right up until the second quarter of 2021, there is very little difference between the best and worst performing models, but it is a dramatically different situation by the third quarter of 2022.
The dense0 model shows a strong difference compared to the other models, except for the second quarter of 2022, when it is outperformed by the xGBRegressor and gru1. However, both of these models fall in accuracy again for the next quarter. 
Looking at Figure \ref{fig:rmaechange}, while there has been a rise in rMAE values, and there is the same separation that is in the third quarter of 2022, the rise has not matched the relative rise in MAE values. The more complex models are performing poorly in recent months. This does highlight the improvement that is seen in the dense0 model, there is a 20 percent reduction of rMAE values between the xGBRegressor and the dense0.

\begin{figure*}[hbt!]
    \centering
    \includegraphics[width=0.7\linewidth]{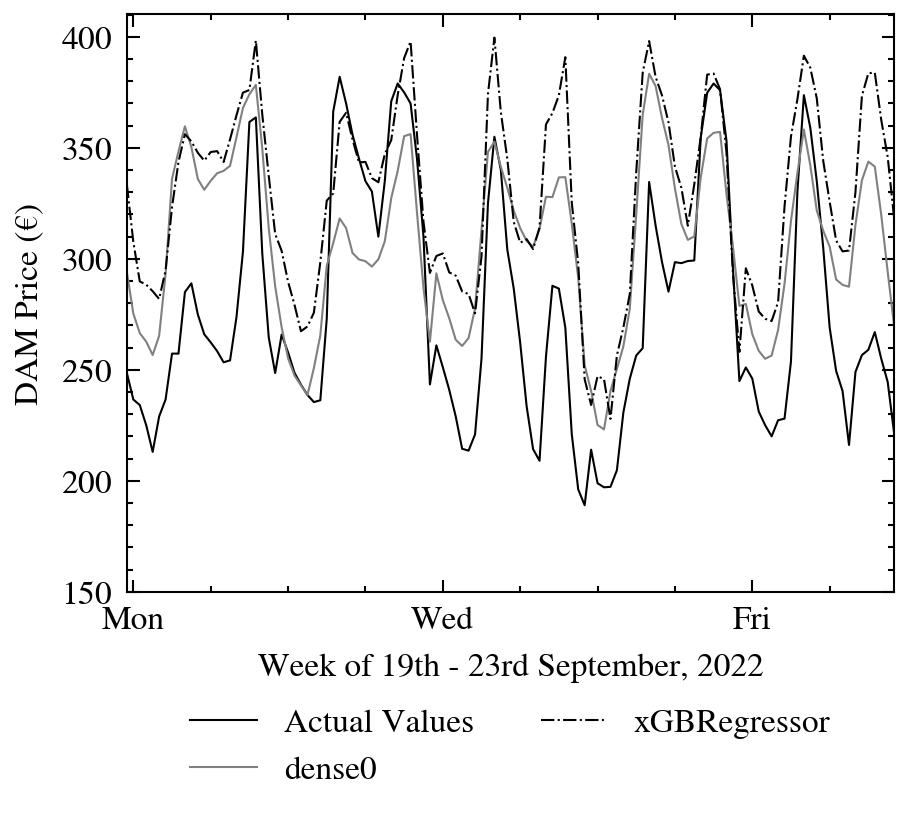}
    \caption{Forcasted DAM price for different models, 19th - 23rd September 2022}
    \label{fig:weekpredict}
\end{figure*}

As can be seen in Figure \ref{fig:weekpredict}, the models follow the rough daily structure will (except for the lower than expected prices on Monday morning and Friday evening). They do regularly overestimate the price, with the xGBRegressor seeming to be particularly affected by this. Of course, this is just a small sample of the test period and the models used.

\section{Conclusion}
In this paper several machine learning and neural network models are analysed on the day-ahead market price for the Integrated Single Electricity Market. The test period is various three month periods, beginning on 1st January 2020 and ending 30th September 2022, with varying lengths of training data.

By analysing how optimised forecasting methods has changed since the beginning of 2020, this paper has demonstrated practical techniques to be used to minimise forecasting errors in periods of tumultuous prices. 
\begin{itemize}

    \item Extreme Gradient Boosted Machines has been shown to be a strong predictor of electricity prices throughout the entire period tested. It is the strongest of all the models in the earlier quarters, pre-April 2021. After this period, a single perceptron with a linear activation function is the most accurate model used. This simple model, when properly trained, outperformed more complex neural network models using convolutional neural networks or recurrent neural networks like long-short term memory and gated recurrent units.
    \item The analysis has demonstrated that one year of historical data provides the best model accuracy. This finding helps to streamline the forecasting process, allowing for faster and more efficient predictions while maintaining a high level of accuracy.
    \item EU Natural Gas prices are more useful for price forecasting for the ISEM, and should be used instead of American Natural Gas prices.
    \item The new metric of Relative Mean Absolute Error can be a useful metric in understanding changes in accuracy in periods of significant price change and volatility.
    \item This study also found that the correlation of features to the day-ahead market price has changed in recent years. Now, the price of natural gas on the day and the amount of wind energy on the grid that hour are significantly more important than any other features. Demand is still more correlated than generation. The relatively small percentage of solar energy on the grid means it is not yet reflected in price correlation values.
\end{itemize}
There can be more research done, on rolling training and testing windows approach, and on models like Deep Extreme Learning Machines \citep{Zhang2022} or ensemble models, like the K-SVM-SVR model that has been research on the Irish markets \citep{Lynch2019} \citep{O'Leary2021} or ensemble models that have worked in other markets, like CNN-LSTM \citep{Xie2018}\citep{Kathirgamanathan2022}\citep{Guo2020} and DNN-GRU/DNN-LSTM \citep{Lago2018}. 
Further research could also look at different structures for the neural networks used. Further testing could be done to find the optimum number of hidden layers and neurons for networks like LSTM, GRU and CNN. Three layers has been shown to work well for GRUs \citep{Ugurlu2018}, as well as the structures listed in the ensemble methods above. There are also some machine learning models to trail on more recent data, for example, Light Gradient Boosted Machines, which worked well in 2020 data in different research on the I-SEM \citep{Lynch2021}, also, LinearSVR, and perhaps SVR could be looked at more as well. Other regression models like Ridge Regression and Lasso Regression should be analysed.

The implications of these findings extend beyond the realm of academia, as stakeholders in the energy sector, including generators, retailers, regulators, and policymakers, can leverage the insights from this study to improve their decision-making processes and better manage the challenges associated with electricity price volatility.
%\section*{Data Statement}
%Data is sourced from the public domain.

\bibliographystyle{elsarticle-num} 
\bibliography{export}

\end{document}